\documentclass[sigconf]{acmart}
\usepackage{multirow} 
\usepackage{makecell}
\usepackage{hyperref}
\hypersetup{hidelinks,
	colorlinks=true,
	allcolors=black,
	pdfstartview=Fit,
	breaklinks=true}

\AtBeginDocument{%
  \providecommand\BibTeX{{%
    \normalfont B\kern-0.5em{\scshape i\kern-0.25em b}\kern-0.8em\TeX}}}

\acmSubmissionID{3627}

\begin{document}

\title{FoRA: Low-Rank Adaptation Model beyond Multimodal Siamese Network}

\author{Wenying Xie}
\email{wyxie@xidian.edu.cn}
\affiliation{%
  \institution{Xidian University}
  \city{Xi'an}
  \state{Shannxi}
  \country{China}
}

\author{Yusi Zhang}
\authornote{Corresponding author}
\email{23011210731@stu.xidian.edu.cn}
\affiliation{%
  \institution{Xidian University}
  \city{Xi'an}
  \state{Shannxi}
  \country{China}
}

\author{Tianlin Hui}
\email{23011210802@stu.xidian.edu.cn}
\affiliation{%
  \institution{Xidian University}
  \city{Xi'an}
  \state{Shannxi}
  \country{China}
}

\author{Jiaqing Zhang}
\email{jqzhang_2@stu.xidian.edu.cn}
\affiliation{%
  \institution{Xidian University}
  \city{Xi'an}
  \state{Shannxi}
  \country{China}
}

\author{Jie Lei}
\email{jielei@mail.xidian.edu.cn}
\affiliation{%
  \institution{Xidian University}
  \city{Xi'an}
  \state{Shannxi}
  \country{China}
}

\author{Yunsong Li}
\email{ysli@mail.xidian.edu.cn}
\affiliation{%
  \institution{Xidian University}
  \city{Xi'an}
  \state{Shannxi}
  \country{China}
}



\begin{abstract}
  Multimodal object detection offers a promising prospect to facilitate robust detection in various visual conditions. However, existing two-stream backbone networks are challenged by complex fusion and substantial parameter increments. This is primarily due to large data distribution biases of multimodal homogeneous information. In this paper, we propose a novel multimodal object detector, named \textbf{L}ow-rank \textbf{M}odal \textbf{A}daptors (\textbf{LMA}) with a shared backbone. The shared parameters enhance the consistency of homogeneous information, while lightweight modal adaptors focus on modality unique features. Furthermore, we design an adaptive rank allocation strategy to adapt to the varying heterogeneity at different feature levels. When applied to two multimodal object detection datasets, experiments validate the effectiveness of our method. Notably, on DroneVehicle, LMA attains a 10.4\% accuracy improvement over the state-of-the-art method with a 149M-parameters reduction. The code is available at \href{https://github.com/zyszxhy/FoRA}{FoRA}.
\end{abstract}





\maketitle

\section{Introduction}
\begin{figure}[h]
  \centering
  \includegraphics[width=\linewidth]{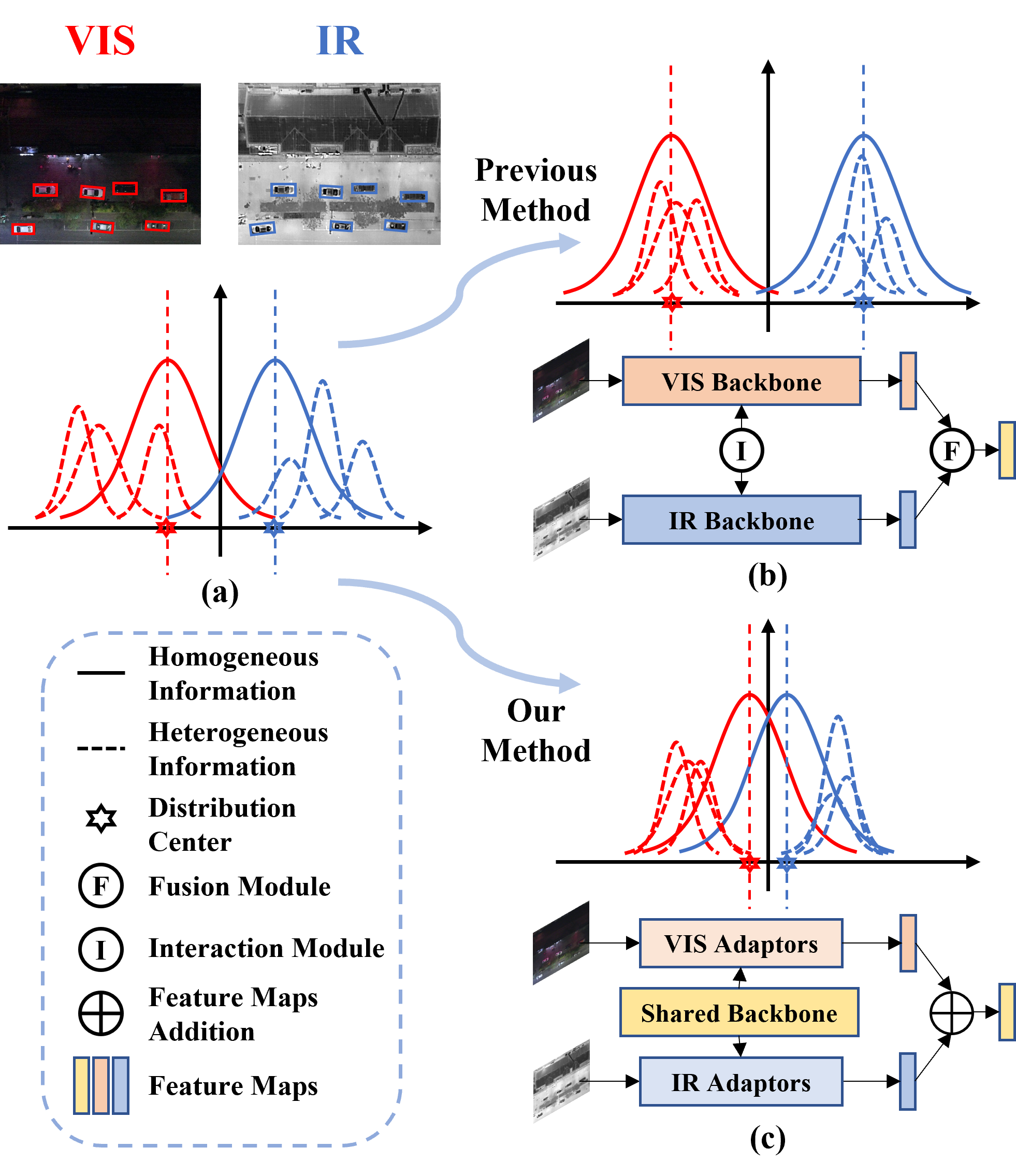}
  \caption{Illustrations of changes in statistical data distributions for valid information and corresponding multimodal object detection structures. (a) Valid information (homogeneous information and heterogeneous information) exhibits different distribution biases in multimodal images. (b) Previous methods employ two-stream backbone (bottom) to extract features, bringing larger distribution biases. (c) We greatly reduce the biases and model scale by combining shared backbone and modal adaptors (bottom).}
  \label{fig1}
  \Description{fig1}
  \vspace{-3mm}
\end{figure}

Object detection is a crucial task in computer vision as it involves locating and identifying interested objects \cite{dynamichead, dynamicdetr}. During the detection, only considering one modality has limitations, e.g. visible images may not provide sufficient high chromatic contrast and visual fidelity information in poor visual and other increasingly complex conditions \cite{sun2022drone,mlpd,hwang2015multispectral,liu1611multispectral}. In contrast, infrared images \cite{huang2203modality,ioannou2014thermal} reflect the surface temperature of objects with temperatures above absolute zero. And they exhibit highlighted contrast information and sharp edge contours, which are less susceptible to changes in visual conditions \cite{kuenzer2013thermal}.

In this case, the assistance of combining visible and infrared images can gain deep insights into multimodal object detection (MOD) to maximize their respective strengths though modality cooperation \cite{qingyun2021cross,qingyun2022cross,zhang2020multispectral,zhang2021guided}. A direct way to perform MOD with deep neural networks (DNNs) is to build a symmetric two-stream backbone where features are extracted independently from each modality. And then these independent multimodal features are fused to complete and enhance the detection performance \cite{qingyun2021cross,chen2023attentive}, as shown in Figure \ref{fig1} (b).

Such an independent two-stream backbone is inadequate due to distribution biases of multimodalities, thus resulting in suboptimal solutions. On the one hand, the two modalities (visible and infrared images) have not been learned together making biases of multimodal features larger. Therefore, well-designed fusion and interaction strategies are need to retain the consistency of homogeneous information, such as Cross-modal Conflict rectification module (CCR) in \cite{he2023multispectral}, Differential Modality Aware Fusion module (DFMA) in \cite{zhou2020improving}, and Inter-modality Cross-Attention module (ICA) in \cite{yuan2024c}. On the other hand, significant and substantial increases in the parameters of additional backbone and fusion module are observed when integrating the unimodal model. For instance, YOLOv8l \cite{Jocher_Ultralytics_YOLO_2023} (44.48M) requires at least one backbone for each additional modality, which results in a parameter increase of 19.78M (44.47\%).

In this paper, we propose a novel multimodal object detector named Low-rank Modal Adaptor (LMA) to address distribution biases of multimodalities while emphasizing lightweight. Different from the previous methods which extract multimodal respective information with \textbf{completely independent network parameters}, LMA introduces \textbf{identical network parameters to extract homogeneous information and distinct parameters to focus on heterogeneous information}. Concretely, we employ a shared backbone to extract the modality-shared features while lightweight modal adaptors for characteristics unique to each modality. Due to the shared parameters, the identification of objects is more uniform because the data distribution bias of homo-information is greatly reduced as illustrated in Figure \ref{fig1}, thus avoiding complex design of fusion and interaction modules. Considering that varying heterogeneity in multimodal features at different network levels seriously affects the choice of parameter counts in adaptors, we employ singular value decomposition (SVD)-like low-rank matrices as our adaptors and design an importance-aware training strategy to dynamically allocate the matrix ranks. Note that adaptors with low-rank matrices own few parameters, which means the network solely introduces little computational overhead for each additional modality. Our contributions can be summarized as follows:
\begin{itemize}
    \item We propose a novel detector named LMA to tackle the issue of large data distribution bias between two modalities. To our best knowledge, our structure is the first work that combines a shared backbone and lightweight modal adaptors for multimodalities, enabling more effective extraction of both homogeneous and heterogeneous information.
    \item To meet the demand for varying parameter counts of adaptors across different multimodal feature levels, we design an adaptive matrix rank allocation strategy that incorporates joint gradients and singular values as importance metric.
    \item We conduct experiments on two multimodal object detection benchmarks: DroneVehicle \cite{sun2022drone} and LLVIP \cite{jia2021llvip}. We prove the large data distribution bias of multimodal features and the varying heterogeneity of different feature levels in two-stream backbone networks. Moreover, our method achieves SOTA and a 10.4\% increase in accuracy over previous methods with a 1000$\times$ reduction in parameter incremental percentages on DroneVehicle dataset. 
\end{itemize}

\section{Related Works}
\subsection{Visible Object Detection}
\textbf{Object Detection} on visible images is a pivotal and extensively utilized task in computer vision. With the publication of numerous benchmark datasets such as Pascal VOC \cite{everingham2015pascal}, ADE20K \cite{zhou2017scene} and COCO-Stuff \cite{caesar2018coco}, object detectors based on deep neural network have been extensively researched. The detection models can be roughly divided into single-stage detectors and two-stage detectors. Two-stage detectors, such as Fast R-CNN \cite{girshick2015fast}, Faster R-CNN \cite{ren2016faster}, can achieve high precision through coarse localization and fine feature extraction of region of interest (RoI), while the model sizes are generally large. In contrast, the structure of the single-stage detectors, such as RetinaNet \cite{lin2017focal}, SSD \cite{liu2016ssd} and YOLO-series detectors \cite{bochkovskiy2020yolov4,chen2021you,chen2021disentangle,Jocher_YOLOv5_by_Ultralytics_2020,ge2021yolox,redmon2016you,redmon2017yolo9000,redmon2018yolov3}, are more lightweight but less accurate. In these methods, Convolutional Neural Networks (CNN) are typically used to construct the backbones. Among various object detection algorithms, single-stage detector YOLOv8 \cite{Jocher_Ultralytics_YOLO_2023} strikes an excellent balance between accuracy and the count of model parameters. Therefore, we have chosen it as our unimodal model.

\textbf{Oriented Object Detection} refers to using rotated bounding boxes to represent detection results, which can provide more accurate location of objects. Detectors based on horizontal bounding box are plagued by multiple objects of interest in one anchor. Therefore, many researches about oriented object detection \cite{ma2018arbitrary,jiang2017r2cnn,ding2019learning,shi2020orientation,zhu2020adaptive} have been presented to alleviate this problem. Oriented R-CNN \cite{xie2021oriented} adjusts receptive fields of neurons in accordance with object shapes and orientations. S2Anet \cite{han2021align} adopts alignments between horizontal receptive fields and rotated anchors to get a better feature representation. In this paper, we replace the detection head of YOLOv8 with an oriented bounding box (OBB) detection head to construct an oriented object detector.

\subsection{Multimodal Object Detection}
Multimodal object detection based on visible-infrared image pairs has become a promising research field to address the issue that visible images are insufficient for capturing crucial information in poor visual conditions \cite{manfredi2020shift,ming2012dynamic,zhou2021ecffnet}. Current multimodal object detection algorithms concentrate on two primary concerns: (1) how to extract valid information of each modality? and (2) how to effectively fuse the multimodal features to generate fusion features that contain distinct object features and complementary information from both modalities? Zhao $et\ al.$ \cite{zhao2024removal} coarsely remove interfering information within each modality and design a dynamic feature selection module to finely select desired features for feature fusion. He $et\ al.$ \cite{he2023multispectral} exam contextual information of analogous pixels to alleviate multimodal information with semantic conflicts, and then fuse multimodal features by assessing intra-modal importance to select semantically rich features and mining inter-modal complementary information. Chen $et\ al.$ \cite{chen2023attentive} emphasize object region while aligning the object regions and utilize a channel-wise attention mechanism with illumination guidance to eliminate the imbalance between different modalities. Zhou $et\ al.$ \cite{zhou2020improving} design a novel Differential Modality Aware Fusion (DMAF) module to make two modalities complement each other and align them adaptively through an illumination aware feature alignment module.

Nonetheless, the aforementioned methods employ two-stream backbone networks to extract features of each modality. To ensure consistency of modality-shared features and heterogeneity of modality-unique features, complex feature interaction and fusion modules have been designed within the backbones. This is challenging in terms of model design and introduces a significant number of additional parameters based on unimodal models. To address this problem, we propose LMA, which utilizes lightweight model structures to enhance the consistency of homogeneous information while preserving heterogeneous information.

\section{Method}
\begin{figure*}[h]
  \centering
  \includegraphics[width=\linewidth]{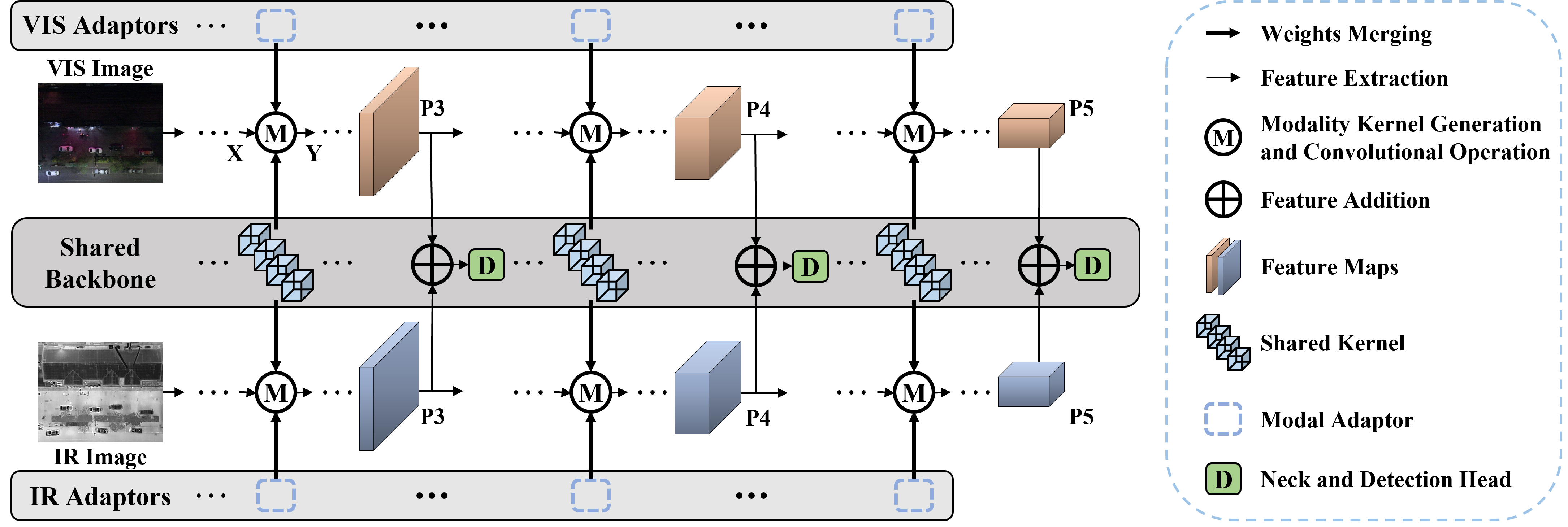}
  \caption{Overview of LMA (Taking convolutional backbone as an example). When a convolutional layer works, a new modality convolutional kernel is first generated by merging the modal adaptor weights and the shared convolutional kernel weights. Then the input data are convolved with the new modality kernel to output the extracted feature map. The details of this process are shown in Figure \ref{fig3} and Figure \ref{fig4}. “P3”, “P4” and “P5” represent the features need to be fused and then fed into detector.}
  \label{fig2}
  \Description{fig2}
\end{figure*}

In this section, we firstly introduce a general formulation of multimodal object detection and point out the limitations of symmetric two-stream backbone networks. Secondly, we detail our detector combining a shared backbone with modal adaptors. We demonstrate the validity of this structure through formulas. Lastly, as our method involves the choices of adaptor parameter counts, we introduce an importance-aware adaptive rank allocation strategy to control the parameter budget.

\subsection{MOD Formulation}
\subsubsection{Formulation of MOD}
As shown in Figure \ref{fig1} (a), assume the distribution of each valid information in multimodal images is Gaussian Distribution with a mean $\mu$ and a variance $\sigma^{2}$, denoted as $X \sim N ( \mu, \sigma^{2} )$. The data distribution of every visible-infrared image pair can be represented as follows:
\begin{equation} \label{eq1}
\begin{gathered}
    D_{mod} \sim C_{mod} N ( \mu_{mod}, \sigma_{mod} {}^{2} )+ \sum_{i=1}^{n_{mod}} C_{i} N ( \mu_{i}, \sigma_{i} {}^{2} ) ,\\
    m o d=\{v, t \},
\end{gathered}
\end{equation} 
where the subscripts $v$ and $t$ refer to the visible and infrared modality respectively. $C_{mod}$ denotes the intensities of homogeneous information in each modality, while $\mu_{mod}$ and ${\sigma_{mod}}^{2}$ are their corresponding mean and variance. Similarly, $C_{i}$ and $\mu_{i}$, ${\sigma_{i}}^{2}$ for $i=1, ..., \ n_{mod}$ represent the intensities of unique information in each modality and their data distribution properties.

In the multimodal object detection, it is desired to generate features that maintain consistency of homogeneous information while minimizing loss of modality-unique information as possible. Therefore, fusion feature maps with ideal data distribution can be represented as follows:
\begin{equation} \label{eq2}
\begin{gathered}
    D_{f} \sim( C_{v}+C_{t} ) N \bigl( \mu_{f}, \sigma_{f} {}^{2} \bigr)+\sum_{i=1}^{n_{v}+n_{t}} C_{i} N ( \mu_{i}, \sigma_{i} {}^{2} ), \\
    s. t. \ \frac{1} {n_{v}+n_{t}} \sum_{i=1}^{( n_{v}+n_{t} )} \mu_{i} \approx\mu_{f} ,
\end{gathered}
\end{equation}
where the merged distribution sums the original intensities of the homogeneous information and generates a new distribution $N \left( \mu_{f}, \sigma_{f} {}^{2} \right) $. Meanwhile, all heterogeneous information intensities are preserved to leverage multimodal complementary information. Ideally, the geometric center of their means is approximated as the mean of the homogeneous information, as shown in Formula (\ref{eq2}).

\subsubsection{Two-stream Backbone Networks}
Deep neural networks (DNN) can be viewed as functions with learnable parameters that maps input data to a high-dimensional feature space. Therefore, the process of feature extraction in two-stream backbone networks can be represented as follows:
\begin{equation} \label{eq3}
    \begin{split}
        E_{mod}&=B ( D_{mod} ; \omega_{mod} ) \\
        &\sim C_{mod}^{*} N \big( \mu_{mod}^{*}, \sigma_{mod}^{*} {}^{2} \big)+\sum_{i=1}^{n_{mod}} C_{i}^{*} N \big( \mu_{i}^{*}, \sigma_{i}^{*} {}^{2} \big), \\
        &\frac{1} {n_{mod}} \sum_{i=1}^{n_{mod}} \mu_{i}^{*} \approx\mu_{r}^{*} .
    \end{split}
\end{equation}
Here, $E$ is the data distribution of modality features and $B$ denotes the mapping function of the backbone with parameters $\omega_{mod}$. The distribution of input data significantly impacts the training and formation of parameters. In turn, a backbone with specific parameters maps the inputs to corresponding data space. As illustrated in Formula (\ref{eq3}) and Figure \ref{fig1} (b), multimodal features extracted by two-stream backbone will form respective new geometric center $\mu_{mod}^{*}$. This will significantly large the biases of homogeneous information and we prove it in subsequent experiments (as illustrated in Figure \ref{fig5}):
\begin{equation} \label{eq4}
    | \mu_{v}^{*}-\mu_{t}^{*} | > | \mu_{v}-\mu_{t} | .
\end{equation}
To obtain the ideal fusion features as shown Formula (\ref{eq2}) from Formula (\ref{eq3}), it is necessary to design efficient fusion functions, which are generally complex and sometimes lack generalization to various datasets. Moreover, two-stream backbones employs completely independent parameters $\omega_{v}$ and $\omega_{t}$ to extract homogeneous information. Such structures are empirically parameter redundant.

\subsection{Proposed Detector LMA}
\subsubsection{Overview of the Framework} \label{sec3.2.1}
The overall architecture of the proposed method is shown in Figure \ref{fig2}. The LMA extends a unimodal model and attaches two lightweight modal adaptors to all feature extraction layers, such as Convolutional, Transformer \cite{vaswani2017attention}, and Fully Connected layers \cite{lecun1989generalization}, of the backbone. Adaptor parameters can be merged with corresponding backbone layer to form a new feature extraction layer with the same shape. Details of modal adapters are introduced in Section \ref{sec3.2.2}.

The parameters of the shared backbone, visible adaptors and infrared adaptors are denoted as $\omega_{shared}$, $\omega_{vada}$ and $\omega_{tada}$ respectively. When a modality data is received, a new backbone layer is created by merging shared layer and corresponding modal adaptor to extract modality features, which can be represented as follows:
\begin{equation} \label{eq5}
    E_{v}=B ( D_{v} ; \omega_{v a d a}+\omega_{s h a r e d} ) ,\quad E_{t}=B ( D_{t} ; \omega_{t a d a}+\omega_{s h a r e d} ) ,
\end{equation}
where shared backbone parameters $\omega_{shared}$ are influenced by both modalities. When acting in turn on multimodal data, they thus tend to map homogeneous information in different modality to the same feature space. The modal adaptors then focus more on the remaining heterogeneous information:
\begin{equation} \label{eq6}
    E_{mod} \sim C^{\prime} {}_{mod} N ( {\mu^{\prime}}_{mod}, {\sigma^{\prime}}^{2} )+\sum_{i=1}^{n_{mod}} {C^{\prime}}_{i} N \big( {\mu^{\prime}}_{i}, {\sigma^{\prime}}_{i}^{2} \big) ,
\end{equation}
\begin{equation} \label{eq7}
    | {\mu^{\prime}}_{v}-{\mu^{\prime}}_{t} | \ll| \mu_{v}^{*}-\mu_{t}^{*} |.
\end{equation}
Formulas (\ref{eq6})-(\ref{eq7}) demonstrate that the shared parameter structure effectively reduces distribution biases in homogeneous information during feature extraction, as shown in Figure \ref{fig1} (c). We also experimentally prove this in subsequent experiments that illustrated in Figure \ref{fig5}. Based on Formulas (\ref{eq6})-(\ref{eq7}), we can obtain the ideal fusion features data distribution approximating that in Formula (\ref{eq2}) through simple addition operation. The method thus gets rid of designs of complex feature fusion mapping functions and enhances robustness of the model.

In terms of model scale, the lightweight adaptors introduce only a small increase in parameters to the unimodal model:
\begin{equation} \label{eq8}
    S ( \omega_{v} )=S ( \omega_{t} )=S ( \omega_{s h a r e d} ) ,
\end{equation}
\begin{equation} \label{eq9}
    and\ S ( \omega_{v a d a} )=S ( \omega_{t a d a} ) \ll S ( \omega_{s h a r e d} ),
\end{equation}
\begin{equation} \label{eq10}
\begin{aligned}
    so\  S(LMA)=S ( \omega_{v a d a} )+S ( \omega_{t a d a} )+S ( \omega_{s h a r e d} ) \\
    \ll S ( \omega_{v} )+S ( \omega_{t} )=S(two-stream\ backbone),
\end{aligned}
\end{equation}
where $S ( \cdot)$ denotes the count of parameters. Formula (\ref{eq10}) means that our method greatly reduces parameter redundancy in multimodal models. In this case, if a new modality $m$ needs to be added, only $S(\omega_{mada})$ will be introduced to the original model. This is a little addition compared to the computational overhead of the entire model.

\subsubsection{Modal Adaptor} \label{sec3.2.2}
\begin{figure}[t]
  \centering
  \includegraphics[width=\linewidth]{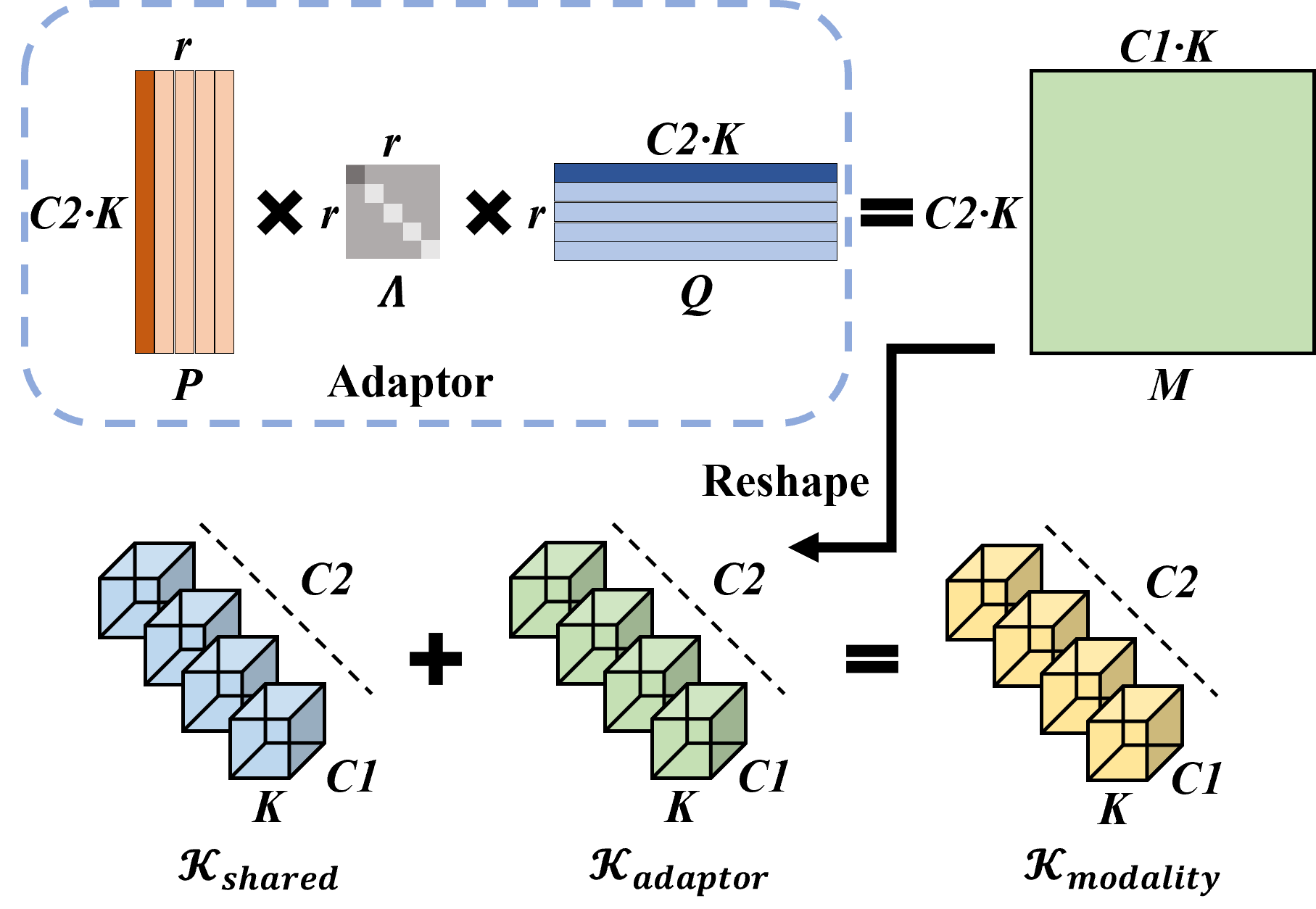}
  \caption{Details of adaptor structure and mergence process of parameters (convolutional layer for example). Adaptors comprise three low-rank matrices $P$, $\Lambda$, $Q$, which generate the adaptor kernel by matrix multiplication and reshaping. Weights of modality kernel are derived from the sum of the shared layer weights and the adaptor kernel weights.}
  \label{fig3}
  \Description{fig3}
\end{figure}

As mentioned in Section \ref{sec3.2.1}, modal adaptors should be lightweight compared to the entire modal and its parameters could be merged with the feature extraction layers of the backbone. Inspired by \cite{hu2021lora} and \cite{zhang2023adaptive}, we apply SVD-like low-rank matrices as modal adaptors. Concretely, taking a convolutional layer of the backbone as an example, the weight parameter, denoted by ${\mathcal K}_{s h a r e d} \in\mathbb{R}^{C 1 \times C 2 \times K \times K}$, is a 4-dimensional tensor, where $K$ represent the width and height of the spatial window of the convolutional kernel, while $C1$ and $C2$ denote the number of input channels of the input to the layer and the number of kernels learned in the layer, respectively. As shown in Figure \ref{fig3}. the adaptor of this layer comprises three low-rank matrices that are multiplied to obtain the adaptor parameter matrix $M$:
\begin{equation} \label{eq11}
    M=P \varLambda Q ,
\end{equation}
where $P \in\mathbb{R}^{C 2 \cdot K \times r} $ and $Q \in\mathbb{R}^{r \times C 1 \cdot K}$ represent the lift and right singular vectors of $M \in\mathbb{R}^{C 2 \cdot K \times C 1 \cdot K}$ and the diagonal matrix $\varLambda\in\mathbb{R}^{r \times r}$ contains the singular values $\{\lambda_{i} \}_{1 \leq i \leq r}$ with inner rank:
\begin{equation} \label{eq12}
    r \ll m i n ( C 1 \cdot K, C 2 \cdot K ).
\end{equation}
We further denote ${\mathcal G}_{i}=\{P_{i}, \lambda_{i}, Q_{i} \}$ as triplet containing the $i \mathrm{-t h} $ singular value and vectors (as shown in Figure \ref{fig3}, the dark part of $P$, $\Lambda$, $Q$ represents a triplet). In practice, since $\Lambda$ is diagonal, we only need to save it as a vector in $\mathbb{R}^{r}$.

\begin{figure}[t]
  \centering
  \includegraphics[width=0.7\linewidth]{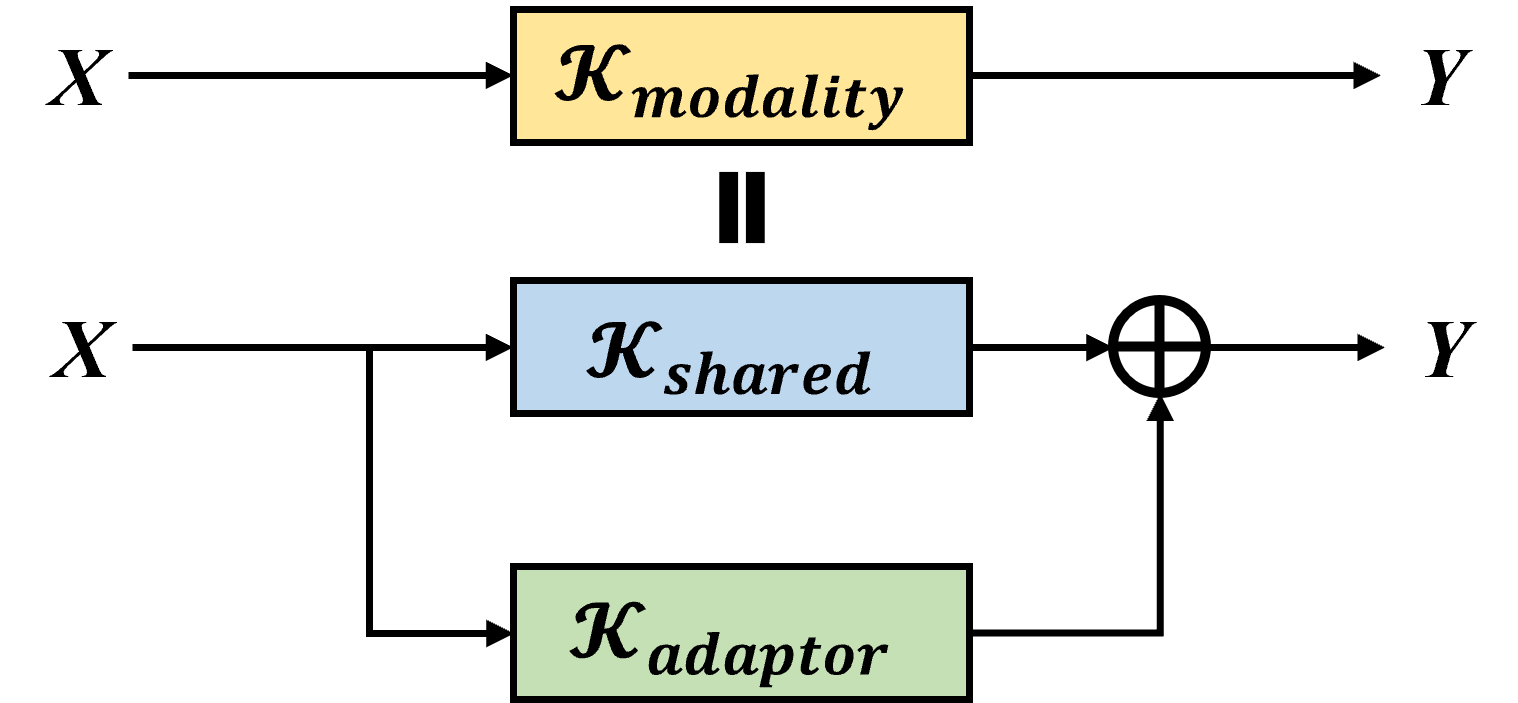}
  \caption{The equivalent data flow process for convolution of input data $X$ and ${\mathcal K}_{m o d a l i l t y}$. $X$ convolves with ${\mathcal K}_{s h a r e d}$ (top branch) and ${\mathcal K}_{a d a p t o r}$ (bottom branch) and results in output $Y$ after summation.}
  \label{fig4}
  \Description{fig4}
\end{figure}

To merge parameters of shared backbone layer and adaptor, we reshape the adaptor matrix to an adaptor convolution kernel $\mathcal{K}_{a d a p t o r} \in\mathbb{R}^{C 1 \times C 2 \times K \times K} $ with the same shape as ${\mathcal K}_{s h a r e d} $:
\begin{equation} \label{eq13}
    \mathcal{K}_{m o d a l i l t y}=\mathcal{K}_{s h a r e d}+\mathcal{K}_{a d a p t o r},
\end{equation}
subsequently, the convolutional kernel ${\mathcal K}_{m o d a l i l t y} \in\mathbb{R}^{C 1 \times C 2 \times K \times K}$ for modal feature extraction is obtained by adding the weights of shared kernel and the adaptor kernel. Following basic theory of convolution \cite{krizhevsky2017imagenet}, modality kernel is equal to the parallel architectural branches illustrated in Figure \ref{fig4}, which enables the layer to focus on different types of information in the input data simultaneously. In addition to convolution layers, adaptor parameter matrices can be reshaped to the shape of various common feature extraction structures. Our method thus could be generalized to any models with common layers.

The calculations of parameter counts for the adaptor and the convolutional layer are as follows:
\begin{equation} \label{eq14}
\begin{aligned}
    S ( a d a p t o r )&=S ( P )+S ( \varLambda )+S ( Q )\\
    &=C 2 \cdot K \cdot r+r \cdot C 1 \cdot K+r\\
    &=r ( K ( C 1+C 2 )+1 ),
\end{aligned}
\end{equation}
\begin{equation} \label{eq15}
    S ( \mathcal{K}_{s h a r e d} )=C 1 \cdot C 2 \cdot K \cdot K ,
\end{equation}
\begin{equation} \label{eq16}
    S ( a d a p t o r ) \leq S ( \mathcal{K}_{s h a r e d} ),
\end{equation}
\begin{equation} \label{eq17}
    when\  r \, \leq\, \frac{C 1 \cdot C 2 \cdot K \cdot K} {K ( C 1+C 2 )+1} \approx\frac{C 1 \cdot C 2 \cdot K} {C 1+C 2}.
\end{equation}
Based on Formula (\ref{eq12}) and Formulas (\ref{eq14})-(\ref{eq17}), Formula (\ref{eq9}) could be easily satisfied.

\subsection{Adaptive Rank Allocation Strategy}
As mentioned in Section \ref{sec3.2.1}, adaptors are designed for heterogeneous information. In DNNs, shallow features primarily contain detailed information such as texture and contours of images, while deep features mainly include the semantic information of the objects. Therefore, the amount of heterogeneous information in features maps between modalities varies with feature levels. We prove this in subsequent experiments as shown in Figure \ref{fig7}. The importance of adaptor matrices thus varies significantly across blocks and layers. Adding more trainable parameters to the critical adaptor matrices can lead to better model performance. In contrast, adding more parameters to those less important matrices yields marginal gains or even hurts model performance. The parameters of the adaptor depend on the inner rank, which is the number of triplets ${\mathcal G}_{i}$ in the SVD matrix. Setting ranks evenly to all adaptor matrices/layers thus often leads to suboptimal performance.

To address this problem, we propose a training strategy to adaptively allocate rank of adaptor based on importance of SVD matrix triplets under limited parameter budget. We define parameter budget $b^{( i t )} $ as the total inner rank of all adaptors, where $i t=0, \ldots, T$ represents the index of training step. We set the initial rank of each adaptor as $r=b^{( 0 )} / n$, where n is the number of adaptors and gradually reduce $b^{( i t )} $ to target budget $b^{( T )} $ during training. We iteratively prune singular values by setting the singular value in triplet of low importance to zero, which will discard the entire triplet, to achieve corresponding budget.

For clear reference, we use $k$ to index the adaptor, i.e., $M_{k}=P_{k} \varLambda_{k} Q_{k}$ for $k=1, \ldots, n$ and denote the $i \mathrm{-t h}$ triplet of $M_{k}$ as ${\mathcal G}_{k, i}=\{P_{k, * i}, \varLambda_{k, i}, Q_{k, i *} \}$ and its importance score as $I S_{k, i}$. Previous methods \cite{cai2010singular,koltchinskii2011nuclear} apply the magnitude of singular values to quantify the importance of every triplet or use the gradient values of the training loss to quantify the importance of every parameter \cite{molchanov2019importance,sanh2020movement,liang2021super,zhang2022platon}. We combine the both metrics to determine the basic importance $I ( \cdot)$ of each entry in the adaptor triplets:
\begin{equation} \label{eq18}
    I ( P_{k, i j} )=I ( Q_{k, i j} )=I ( \omega_{k, i j} )=| \nabla_{\omega_{k, i j}} {\mathcal{L}} ( P, \varLambda, Q ) | ,
\end{equation}
\begin{equation} \label{eq19}
    I ( \varLambda_{k, i} )=| \varLambda_{k, i} \cdot\nabla_{\varLambda_{k, i}} {\mathcal{L}} ( P, \varLambda, Q ) |,
\end{equation}
where $\omega_{k, i j} $ is any trainable parameter in lift and right singular vectors. $\mathcal{L} ( \cdot)$ is the loss of the task, which contains categorical loss and regression loss in object detection, and $\nabla$ denotes the gradient operation. Here, importance of singular vectors is determined only by gradients of loss, while that of singular vales is quantified by the magnitude of gradient-weight product. However, such a score in Equations (\ref{eq18})-(\ref{eq19}) is estimated on the sampled mini batch. The stochastic sampling and complicated training dynamics incur high variability and large uncertainty for estimating the importance. Therefore, we utilize the importance smoothing and uncertainty quantification in \cite{zhang2022platon} to resolve this issue:
\begin{equation} \label{eq20}
    \bar{I}^{( i t )}=\beta_{1} \bar{I}^{( i t-1 )}+( 1-\beta_{1} ) I^{( i t )} ,
\end{equation}
\begin{equation} \label{eq21}
    \overline{{{{U}}}}^{( i t )}=\beta_{2} \overline{{{{U}}}}^{( i t-1 )}+( 1-\beta_{2} ) | I^{( i t )}-\bar{I}^{( i t )} | ,
\end{equation}
where $0 < \beta_{1}, \beta_{2} < 1$. $\bar{I}^{( i t )} $ is the smoothed importance by exponential moving average and ${\bar{U}}^{( i t )}$ is the uncertainty term quantified by the local variation between $I^{( i t )}$ and $\bar{I}^{( i t )} $. Then the importance score of each entry is defined as the product between $\bar{I}^{( i t )} $ and ${\bar{U}}^{( i t )}$ and denoted as $s ( \cdot) $:
\begin{equation} \label{eq22}
    s^{( i t )}=\bar{I}^{( i t )} \cdot\bar{U}^{( i t )} .
\end{equation}
The importance score of $i \mathrm{-t h}$ triplet of $M_{k}$ can be represented as:
\begin{equation} \label{eq23}
    I S_{k, i}=s ( \varLambda_{k, i} )+\frac{1} {C 2 \cdot K} \sum_{j=1}^{C 2 \cdot K} s ( P_{k, j i} )+\frac{1} {C 1 \cdot K} \sum_{j=1}^{C 1 \cdot K} s ( Q_{k, i j} ) ,
\end{equation}
where we employ the mean importance of $P_{k, * i}$ and $Q_{k, i *} $ as its importance score such that $I S_{k, i}$ does not scale with the number of parameters in triplet.

All trainable parameters of adaptors can be denoted as $P=\{P_{k} \}_{k=1}^{n}$, ${\varLambda}=\{{\varLambda}_{k} \}_{k=1}^{n}$ and $Q=\{Q_{k} \}_{k=1}^{n}$. At the $i \mathrm{-t h}$ step, we first update the trainable parameter by gradient descent:
\begin{equation} \label{eq24}
\begin{aligned}
    P_{k}^{( i t+1 )}&=P_{k}^{( i t )}-\eta\nabla_{P_{k}} \mathcal{L} ( P, \Lambda, Q ) ; \; \\
    Q_{k}^{( i t+1 )}&=Q_{k}^{( i t )}-\eta\nabla_{Q_{k}} \mathcal{L} ( P, \Lambda, Q ) ,
\end{aligned}
\end{equation}
\begin{equation} \label{eq25}
    \tilde{\varLambda}_{k}^{( i t )}=\varLambda_{k}^{( t )}-\eta\nabla_{\varLambda_{k}} \mathcal{L} ( P, \varLambda, Q ) ,
\end{equation}
where $\eta> 0 $ is learning rate. Then, given importance score of triplet set $I S_{k}^{( i t )}$ that contains all adaptor martrices, the singular values are pruned following:
\begin{equation} \label{eq26}
\Lambda_{k}^{( i t+1 )}=
    \begin{cases}
        {\widetilde{\Lambda}}_{k, i}^{( i t )},  & \text{if $I S_{k, i}^{( i t )}$ is in the $t o p-b^{( i t )}$ of $I S_{k}^{( i t )}$ } \\
        0, & \text{otherwise}
    \end{cases}.
\end{equation}
In this way, we adaptively allocate higher rank to more important adaptors under a limited parameter budget.

\section{Experiments}

\begin{table*}
  \caption{Detection results (mAP@0.5, in \%) on DroneVehicle val dateset. "Unimodal Model" represents the base unimodal model of the multimodal detector, the same is true in Table \ref{tab2} and Table \ref{tab3}. Note that all detectors locate and classify vehicles with OBB (Oriented Bounding Box) heads. Best results highlighted in \textbf{BOLD}.}
  \label{tab1}
  \begin{tabular}{c|c|ccccc|c|c}
    \toprule[1.5pt]
    Detectors &Unimodal Model & Car & Truck & Freight Car & Bus & Van & mAP@0.5 & Modal\\
    \midrule[1.5pt]
    RetinaNet \cite{lin2017focal}     & - & 78.5 & 34.4 & 24.1 & 69.8 & 28.8 & 47.1 & \multirow{5}{*}{Visible}\\
    Faster R-CNN \cite{ren2016faster}   & - & 79.7 & 42.0 & 34.0 & 76.9 & 37.7 & 54.1 &\\
    S2ANet \cite{han2021align}        & - & 79.9 & 50.0 & 36.2 & 82.8 & 37.5 & 57.3 &\\
    Oriented R-CNN \cite{xie2021oriented} & - & 80.3 & 55.4 & 42.1 & 86.8 & 46.9 & 62.3 &\\
    YOLOv8l \cite{Jocher_Ultralytics_YOLO_2023}       & - & 97.3 & 81.4 & 65.1 & 97.3 & 68.1 & 81.8 &\\
    \midrule
    RetinaNet \cite{lin2017focal}     & - & 88.8 & 35.4 & 39.5 & 76.5 & 32.1 & 54.5 & \multirow{5}{*}{Infrared}\\
    Faster R-CNN \cite{ren2016faster}  & - & 89.7 & 41.0 & 43.1 & 86.3 & 41.2 & 60.3 &\\
    S2ANet \cite{han2021align}        & - & 89.7 & 51.0 & 50.3 & 89.0 & 44.0 & 64.8 &\\
    Oriented R-CNN \cite{xie2021oriented} & - & 89.6 & 53.9 & 53.9 & 89.2 & 41.0 & 65.5 &\\
    YOLOv8l \cite{Jocher_Ultralytics_YOLO_2023}       & - & 98.5 & 69.5 & 73.8 & \textbf{98.0} & \textbf{72.0} & 82.4 &\\
    \midrule
    Halfway Fusion \cite{liu2016multispectral} & Faster R-CNN   & 89.9 & 60.3 & 55.5 & 89.0 & 46.3 & 68.2 & \multirow{6}{*}{Visible+Infrared}\\
    AR-CNN \cite{zhang2021weakly}        & Faster R-CNN   & 90.1 & 64.8 & 62.1 & 89.4 & 51.5 & 71.6 &\\
    TSFADet \cite{yuan2022translation}       & Oriented R-CNN & 89.9 & 67.9 & 63.7 & 89.8 & 54.0 & 73.1 &\\
    CALNet \cite{he2023multispectral}        & ResNet50       & 90.1 & 70.7 & 67.5 & 89.7 & 55.0 & 74.6 &\\
    CALNet \cite{he2023multispectral}        & YOLOv5l        & 90.3 & 73.7 & 68.7 & 89.7 & 59.7 & 76.4 &\\
    \textbf{LMA}   & YOLOv8l        & \textbf{98.7} & \textbf{85.5} & \textbf{81.2} & 96.8 & 71.7 & \textbf{86.8} &\\
    \bottomrule[1.5pt]
  \end{tabular}
\end{table*}

\subsection{Dataset and Evaluation Metrics}
\textbf{DroneVehicle Dataset} \cite{sun2022drone} is a large-scale drone-based visible-infrared dataset that contains 953,087 vehicle instances in 56,878 images with oriented bounding boxes for five categories (car, bus, truck, van and freight car). The dataset is collected using drone platform under different scenes and lighting conditions. The whole dataset is split into a training set, a validation set and a test set.

\textbf{LLVIP Dataset} \cite{jia2021llvip} is a very recently released visible-infrared paired pedestrians dataset which is collected in low-light environments. And most of the images were captured in very dark scenes. The dataset contains 16,836 image pairs with horizontal boxes for one category (person).

\textbf{Evaluation Metrics.} We employ the COCO-style metric of the mean average precision (mAP) to assess the accuracy of MOD. For mAP, an Intersection over Union (IoU) threshold such as 0.5 (denoted as mAP@0.5), 0.5-0.95 (denoted as mAP) is used to calculate True Positives (TP) and False Positives (FP). Meanwhile, we utilize the parameter counts (Megabyte, denoted as M) to characterize the size of models, along with its computational and storage overheads.

\subsection{Implementation Details}
We select YOLOv8l as the unimodal model. We use oriented detection head for DroneVehicle dataset while horizontal detection head for LLVIP. All detectors are trained for 50 epochs with a batch size of 8 on an NVIDIA A100-SXM-80GB GPU. We optimize the training process using Adam algorithm. The input image size is set to $640\times640$. We set the initial average rank to 9 and the target average rank to 6. We warm up the training for 8 epochs, where the rank budget remains constant, and then follow a cubic schedule to decrease the budget until it reaches target budget in the 25-th epoch. After that, we fix the rank allocation and train the model for last 25 epochs. All experiments are conducted based on the modified code library Ultralytics \cite{Jocher_Ultralytics_YOLO_2023} and all other hyperparameters is the same as its default settings.

\subsection{Comparison with State-of-the-Art Methods}
\textbf{Comparison of Accuracy on DroneVehicle.} We compare unimodal models and our LMA with the recent state-of-the-art methods on the val set of the DroneVehicle (shown in Table \ref{tab1}). Among the single-modal methods, the YOLOv8l outperforms all single-modal detectors in both modalities. Detection with infrared modality generally performs better than detection with visible modality. The YOLOv8l exhibits lower accuracy in detecting ‘Truck’ in infrared images compared to visible images, while the opposite is true for ‘Freight Car’. Among the multi-modal methods, our method utilizes the complementary information from both modalities and achieve the highest accuracy in three categories -- ‘Car’, ‘Truck’ and ‘Freight Car’. Multimodal detector LMA achieves a 4.4\% mAP@0.5 higher than the unimodal model. In addition, \textbf{LMA outperforms the current state-of-the-art method by 10.4\% in mAP@0.5!}

\textbf{Comparison of Accuracy on LLVIP.} As shown in Table \ref{tab2}, our proposed method is compared with several state-of-the-art detectors namely Cascade R-CNN \cite{cai2018cascade}, CFT \cite{qingyun2021cross} and CALNet \cite{he2023multispectral}. LMA achieves promising performance on this dataset, outperforming the highest accuracy among all methods by 1.8\% mAP. The superior performance on both multimodal object detection datasets demonstrates the effectiveness and robustness of LMA.

\begin{table}
  \caption{Detection results (mAP, in \%) on LLVIP dateset. All methods detect with visible modality and Infrared modality. Best results highlighted in BOLD.}
  \label{tab2}
  \begin{tabular}{c|c|c}
    \toprule[1.5pt]
    Detectors     & Unimodal Model & mAP  \\
    \midrule[1.5pt]
    Faster R-CNN \cite{ren2016faster} & Faster R-CNN   & 56.9 \\
    RetinaNet \cite{lin2017focal}    & RetinaNet      & 56.9 \\
    Cascade R-CNN \cite{cai2018cascade} & Cascade R-CNN  & 59.6 \\
    CFT   \cite{qingyun2021cross}        & YOLOv5l        & 63.6 \\
    CALNet \cite{he2023multispectral}  & ResNet50       & 63.4 \\
    CALNet \cite{he2023multispectral}  & YOLOv5l        & 63.9 \\
    \textbf{LMA}  & YOLOv8l        & \textbf{65.7} \\
  \bottomrule[1.5pt]
\end{tabular}
\end{table}

\textbf{Comparison of Network Parameters.} As shown in Table \ref{tab3}, previous methods would introduce over 50\% of the parameter counts to adapt the inputs of two modalities based on unimodal models. Otherwise, the unimodal models are already of large size. Moreover, if there are complex fusion or interaction modules in multimodal detectors, the parameter increments would reach more than three times of the original models. Among the five methods, \textbf{LMA is the smallest detector with only a 0.34\% parameter increment.} We reduce the parameter increment by approximately $1000\times$ compared to the state-of-the-art method while still achieving the best performance. This demonstrates the effectiveness of our method in reducing parameter redundancy.

\begin{table}
  \caption{Parameter analysis of detectors. "Param Increment" demonstrates the incremental parameters of multimodal object detectors based on the unimodal models.}
  \label{tab3}
  \begin{tabular}{c|c|c|c}
    \toprule[1.5pt]
    Detectors      & Params (M) & \makecell[c]{Unimodal \\ Params (M)} & Param Increment    \\
    \midrule[1.5pt]
    Halfway Fusion & 144.86     & 129.62              & 11.76\%(15.24M)   \\
    Faster R-CNN   & 64.81      & 41.53               & 56.06\%(23.28M)   \\
    CFT            & 206.03     & 47.05               & 337.90\%(158.98M) \\
    CALNet         & 193.64     & 47.05               & 311.56\%(146.59M) \\
    \textbf{LMA}   & \textbf{44.63}      & \textbf{44.48}   & \textbf{0.34\%(0.15M)} \\
  \bottomrule[1.5pt]
\end{tabular}
\end{table}

\subsection{Ablation Studies}
Ablation experiments are conducted on the DroneVehicle val set. We introduce the Pearson product-moment correlation coefficient ($| \rho|$) between two modality feature maps to represent the biases of the multimodal information data distribution.

\begin{figure}[t]
  \centering
  \includegraphics[width=\linewidth]{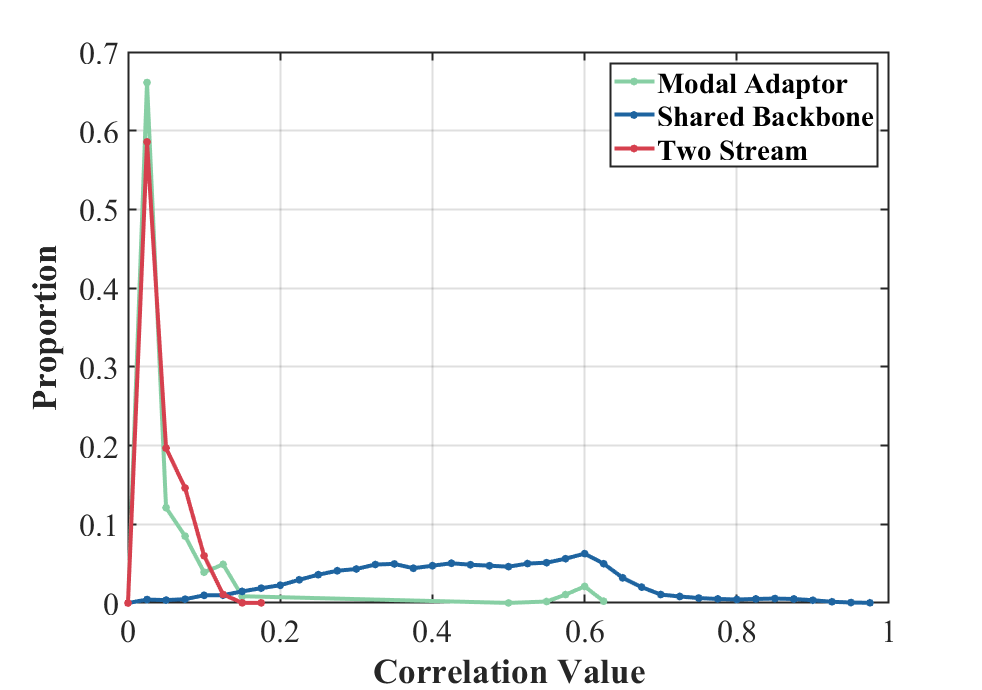}
  \caption{The data distribution biases (Pearson product-moment correlation coefficient ($| \rho|$) between two modality feature maps. The red line represents the baseline feature maps. The green and blue lines represent feature maps extracted by adaptor kernels and shared kernels respectively.}
  \label{fig5}
  \Description{fig5}
\end{figure}

\textbf{Ablation on the Structure of LMA.} The baseline is a two-stream backbone detector based on YOLOv8l and fused by addition. We calculate the Pearson product-moment correlation coefficient between two modality feature maps of multimodal detectors in the stages that correspond to the P3, P4, P5 in YOLOv8l. For LMA, the modal kernels are split into shared kernels and adaptor kernels as shown in Figure \ref{fig4} to extract the homogenous information and heterogeneous information separately. We thus calculate the distribution biases of features generated by
both shared backbone and modal adaptors. The statistics are made on the correlation coefficients $| \rho|$ between two modalities and illustrate the proportion of different correlation coefficients $| \rho|$ ($0. 0 \sim1. 0$) as a line chart in Figure \ref{fig5}. There is a significant bias between the data distributions of all the multimodal information extracted by two-steam backbone networks. In contrast, LMA utilizes a shared backbone to enhance the consistency of homogeneous information while preserving the high variability of multimodal heterogeneous information using modal adaptors.

\begin{figure}[t]
  \centering
  \includegraphics[width=\linewidth]{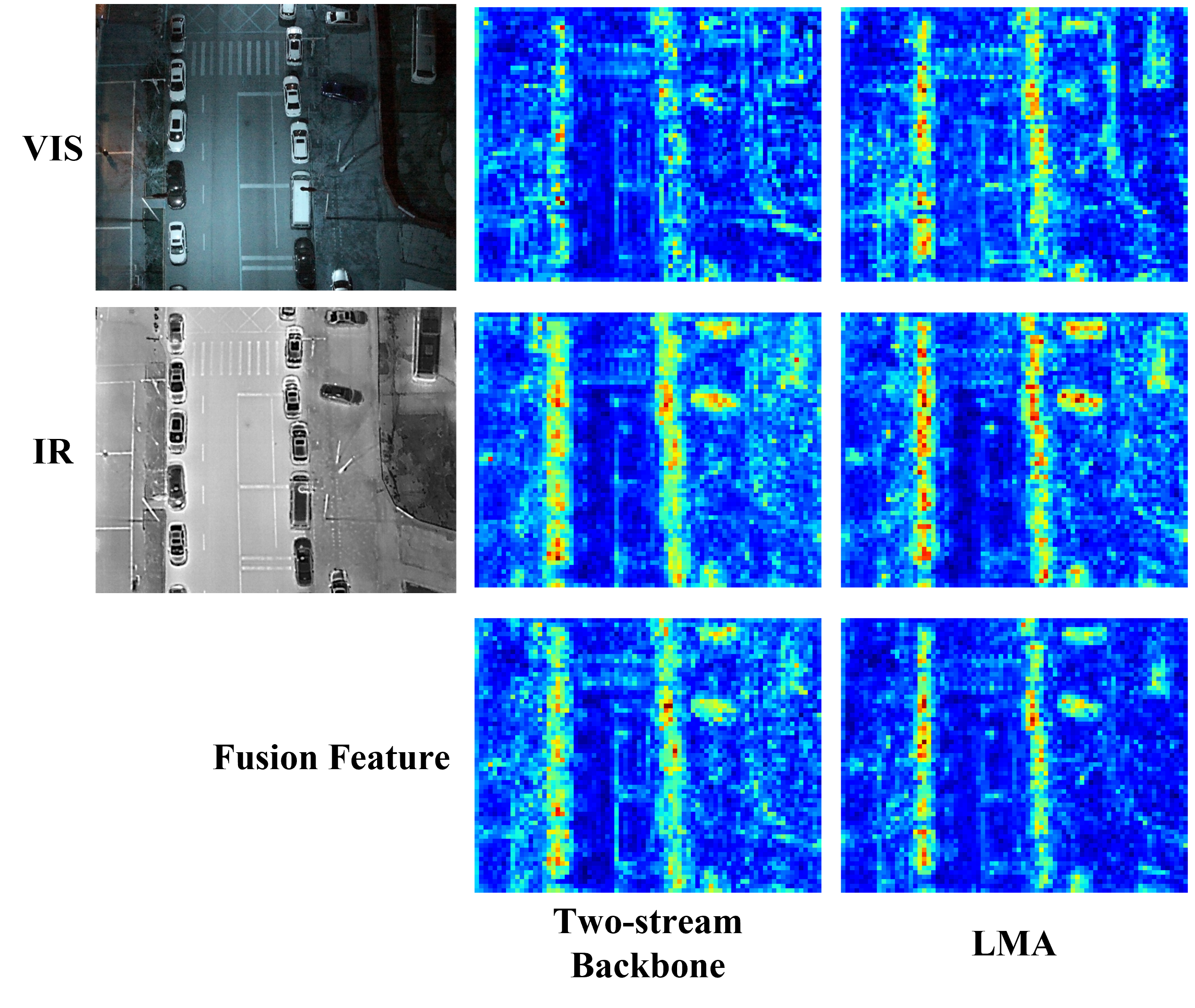}
  \caption{Visualization of feature maps in P3 of LMA and two-stream backbone network.}
  \label{fig6}
  \Description{fig6}
\end{figure}

We compare the accuracy performance and parameters of LMA with the baseline as shown in Table \ref{tab4}. LMA achieves a 1.7\% mAP@0.5 improvement in accuracy while reducing parameter increment by a factor of 130, which means we could introduce other modalities with little concern for parameter limitations. Figure \ref{fig6} presents the visualization results of LMA. In two-stream backbone networks, the data distributions of objects and background information in each modality tend to be similar due to the complete independence of feature extraction. In addition, the difference of data distribution between modalities exacerbates the issue that object features in fusion feature maps are obscured in a large amount of background noise. LMA enhances the representation of homogeneous object features in each modality, while suppressing the extraction of heterogeneous noise to some extent. This makes the object information in the fusion features more significant.

\begin{figure}[t]
  \centering
  \includegraphics[width=\linewidth]{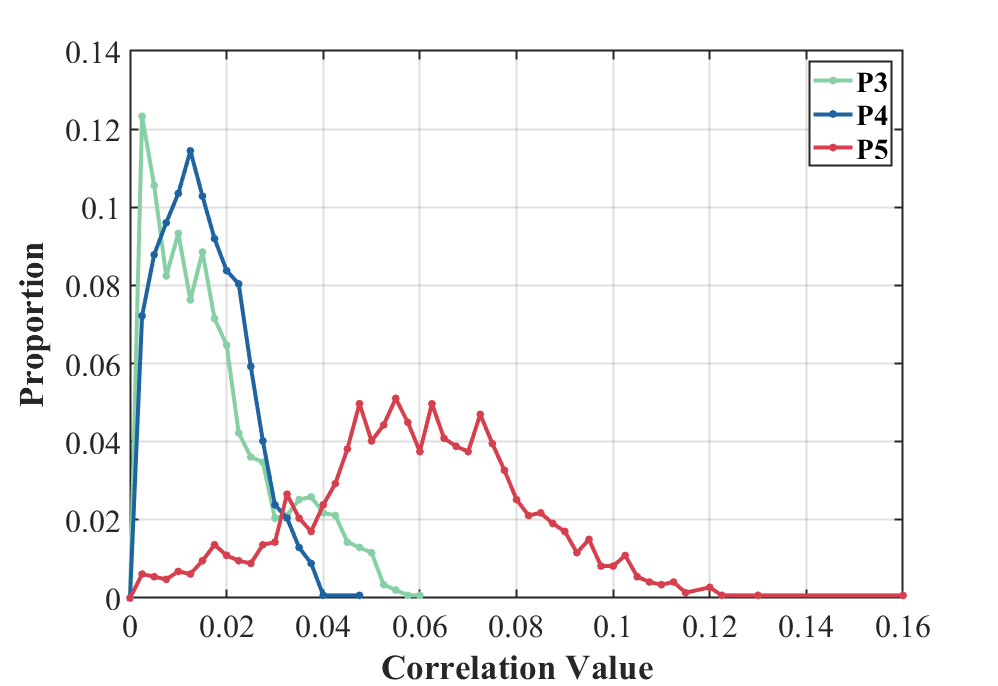}
  \caption{The amount of heterogeneous information (biases of data distributions between multimodal feature maps) in different levels of the network.}
  \label{fig7}
  \Description{fig7}
\end{figure}

\begin{table}
  \caption{Comparison of accuracy and parameters between LMA and two baselines in two ablation experiments on DroneVehicle val set.}
  \label{tab4}
  \begin{tabular}{c|c|c|c}
    \toprule[1.5pt]
    Detectors & mAP@0.5 & Params (M) & Param Increment  \\
    \midrule[1.5pt]
    \makecell[c]{Two-stream \\ Backbone} & 85.1 & 64.26 & 44.47\%(19.78M) \\
    \midrule
    \makecell[c]{LMA with \\ fixed rank of 6} & 86.4    & 45.13 & 1.46\%(0.65M) \\
    \midrule
    \textbf{LMA} & \textbf{86.8}  & \textbf{44.63} & \textbf{0.34\%(0.15M)} \\
  \bottomrule[1.5pt]
\end{tabular}
\end{table}

\textbf{Ablation on the Adaptive Rank Allocation Strategy.} The baseline is a LMA detector with all adaptor matrices have evenly been set fixed rank to 6. Empirically, if there is a high degree of heterogeneity between two modalities, the data distribution of their feature maps will be biased. We thus utilize the multimodal data distribution biases to represent the amount of heterogeneous information. We extracted the features that would be fused in the three stages, P3, P4 and P5, of the two-stream baseline detector. And the data distribution biases between the multimodal feature maps of each stage are calculated separately. As shown in Figure \ref{fig7}, the data distribution bias is significantly lower in the P5 stage compared to the P3 stage. The experimental results demonstrate that shallow features contain more heterogeneous information than deep features in multimodal networks. 

\begin{figure}[t]
  \centering
  \includegraphics[width=\linewidth]{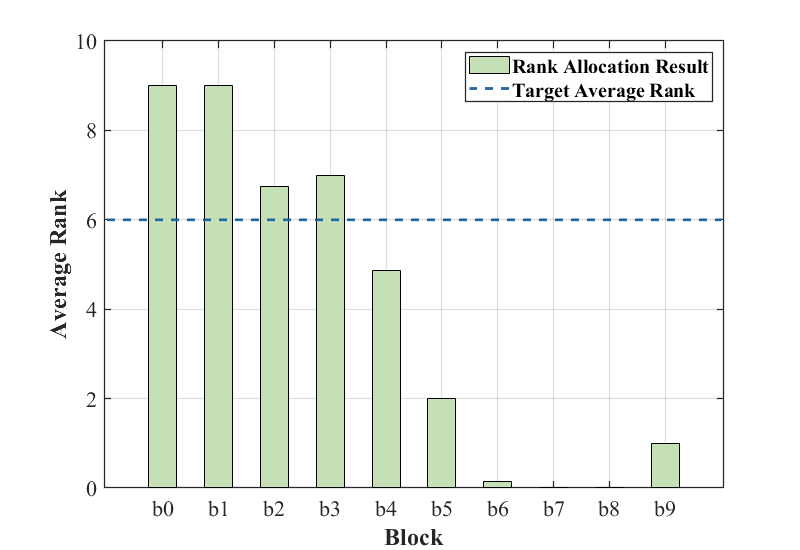}
  \caption{Rank allocation results (average rank of adaptors in each block within the backbone) after adaptive rank allocation. There are total 10 blocks in the backbone of YOLOv8l, denoted as ‘b1’ to ‘b10’ in data flow order.}
  \label{fig8}
  \Description{fig8}
\end{figure}

Figure \ref{fig8} displays the average adaptor rank for each block in the backbone of the LMA following adaptive rank allocation strategy. As analyzed, LMA tends to allocate higher ranks to blocks of shallow levels such as ‘b0’, ‘b1’, ‘b2’ and ‘b3’, while allocating lower ranks to deep levels such as ‘b6’, ‘b7’, ‘b8’ and ‘b9’. The lower-level blocks all have an average rank higher than the target rank, or even the same as the initial rank of the setup. In contrast, the higher-level blocks have smaller averages, some (‘b7’ and ‘b8’) even 0, indicating that the visible and infrared modalities share network parameters in these blocks. The experimental results in Table \ref{tab4} demonstrate that the adaptive rank allocation strategy mitigates parameter redundancy, while improving the accuracy performance of the detector.

\section{Conclusion}
In this paper, we theoretically elaborate on the feature fusion process in multimodal two-stream backbone networks from the perspective of data distribution and point out their limitations. To solve this, we propose a novel multimodal detector integrated a shared backbone and lightweight modal adaptors, named LMA, to mitigate the distribution biases between homogeneous information. In addition, we design an adaptive rank allocation strategy to account for the varying heterogeneity in multimodal features across different levels. Extensive experiments are conducted on two multimodal detection datasets. We experimentally prove the large distribution biases and varying heterogeneity in multimodal features. The experimental results demonstrate that LMA achieves higher accuracy compared to two-stream backbone networks with minimal computation overhead. Specifically, on DroneVehicle val set, LMA achieves 10.4\% accuracy improvement compared to the state-of-the-art method while achieving a 149M-parameters reduction. In the future, we plan to work on how to fuse heterogeneous information effectively with still little computation overhead, which we have not take into account in this paper.

\clearpage

\bibliographystyle{ACM-Reference-Format}
\bibliography{Mybib}

\end{document}


\title{Supplementary Materials: The Name of the Title is Hope}


\author{Anonymous Authors}

\maketitle

\section{Introduction}
ACM's consolidated article template, introduced in 2017, provides a
consistent \LaTeX\ style for use across ACM publications, and
incorporates accessibility and metadata-extraction functionality
necessary for future Digital Library endeavors. Numerous ACM and
SIG-specific \LaTeX\ templates have been examined, and their unique
features incorporated into this single new template.

If you are new to publishing with ACM, this document is a valuable
guide to the process of preparing your work for publication. If you
have published with ACM before, this document provides insight and
instruction into more recent changes to the article template.

The ``\verb|acmart|'' document class can be used to prepare articles
for any ACM publication --- conference or journal, and for any stage
of publication, from review to final ``camera-ready'' copy, to the
author's own version, with {\itshape very} few changes to the source.

\section{Template Overview}
As noted in the introduction, the ``\verb|acmart|'' document class can
be used to prepare many different kinds of documentation --- a
dual-anonymous initial submission of a full-length technical paper, a
two-page SIGGRAPH Emerging Technologies abstract, a ``camera-ready''
journal article, a SIGCHI Extended Abstract, and more --- all by
selecting the appropriate {\itshape template style} and {\itshape
  template parameters}.

This document will explain the major features of the document
class. For further information, the {\itshape \LaTeX\ User's Guide} is
available from
\url{https://www.acm.org/publications/proceedings-template}.

\subsection{Template Styles}

The primary parameter given to the ``\verb|acmart|'' document class is
the {\itshape template style} which corresponds to the kind of publication
or SIG publishing the work. This parameter is enclosed in square
brackets and is a part of the {\verb|documentclass|} command:
\begin{verbatim}
  \documentclass[STYLE]{acmart}
\end{verbatim}

Journals use one of three template styles. All but three ACM journals
use the {\verb|acmsmall|} template style:
\begin{itemize}
\item {\verb|acmsmall|}: The default journal template style.
\item {\verb|acmlarge|}: Used by JOCCH and TAP.
\item {\verb|acmtog|}: Used by TOG.
\end{itemize}

The majority of conference proceedings documentation will use the {\verb|acmconf|} template style.
\begin{itemize}
\item {\verb|acmconf|}: The default proceedings template style.
\item{\verb|sigchi|}: Used for SIGCHI conference articles.
\item{\verb|sigchi-a|}: Used for SIGCHI ``Extended Abstract'' articles.
\item{\verb|sigplan|}: Used for SIGPLAN conference articles.
\end{itemize}

\subsection{Template Parameters}

In addition to specifying the {\itshape template style} to be used in
formatting your work, there are a number of {\itshape template parameters}
which modify some part of the applied template style. A complete list
of these parameters can be found in the {\itshape \LaTeX\ User's Guide.}

Frequently-used parameters, or combinations of parameters, include:
\begin{itemize}
\item {\verb|anonymous,review|}: Suitable for a ``dual-anonymous''
  conference submission. Anonymizes the work and includes line
  numbers. Use with the \verb|\acmSubmissionID| command to print the
  submission's unique ID on each page of the work.
\item{\verb|authorversion|}: Produces a version of the work suitable
  for posting by the author.
\item{\verb|screen|}: Produces colored hyperlinks.
\end{itemize}

This document uses the following string as the first command in the
source file:
\begin{verbatim}
\documentclass[sigconf,authordraft]{acmart}
\end{verbatim}

\section{Modifications}

Modifying the template --- including but not limited to: adjusting
margins, typeface sizes, line spacing, paragraph and list definitions,
and the use of the \verb|\vspace| command to manually adjust the
vertical spacing between elements of your work --- is not allowed.

{\bfseries Your document will be returned to you for revision if
  modifications are discovered.}

\section{Typefaces}

The ``\verb|acmart|'' document class requires the use of the
``Libertine'' typeface family. Your \TeX\ installation should include
this set of packages. Please do not substitute other typefaces. The
``\verb|lmodern|'' and ``\verb|ltimes|'' packages should not be used,
as they will override the built-in typeface families.

\section{Title Information}

The title of your work should use capital letters appropriately -
\url{https://capitalizemytitle.com/} has useful rules for
capitalization. Use the {\verb|title|} command to define the title of
your work. If your work has a subtitle, define it with the
{\verb|subtitle|} command.  Do not insert line breaks in your title.

If your title is lengthy, you must define a short version to be used
in the page headers, to prevent overlapping text. The \verb|title|
command has a ``short title'' parameter:
\begin{verbatim}
  \title[short title]{full title}
\end{verbatim}

\section{Authors and Affiliations}

Each author must be defined separately for accurate metadata
identification. Multiple authors may share one affiliation. Authors'
names should not be abbreviated; use full first names wherever
possible. Include authors' e-mail addresses whenever possible.

Grouping authors' names or e-mail addresses, or providing an ``e-mail
alias,'' as shown below, is not acceptable:
\begin{verbatim}
  \author{Brooke Aster, David Mehldau}
  \email{dave,judy,steve@university.edu}
  \email{firstname.lastname@phillips.org}
\end{verbatim}

The \verb|authornote| and \verb|authornotemark| commands allow a note
to apply to multiple authors --- for example, if the first two authors
of an article contributed equally to the work.

If your author list is lengthy, you must define a shortened version of
the list of authors to be used in the page headers, to prevent
overlapping text. The following command should be placed just after
the last \verb|\author{}| definition:
\begin{verbatim}
  \renewcommand{\shortauthors}{McCartney, et al.}
\end{verbatim}
Omitting this command will force the use of a concatenated list of all
of the authors' names, which may result in overlapping text in the
page headers.

The article template's documentation, available at
\url{https://www.acm.org/publications/proceedings-template}, has a
complete explanation of these commands and tips for their effective
use.

Note that authors' addresses are mandatory for journal articles.

\section{Rights Information}

Authors of any work published by ACM will need to complete a rights
form. Depending on the kind of work, and the rights management choice
made by the author, this may be copyright transfer, permission,
license, or an OA (open access) agreement.

Regardless of the rights management choice, the author will receive a
copy of the completed rights form once it has been submitted. This
form contains \LaTeX\ commands that must be copied into the source
document. When the document source is compiled, these commands and
their parameters add formatted text to several areas of the final
document:
\begin{itemize}
\item the ``ACM Reference Format'' text on the first page.
\item the ``rights management'' text on the first page.
\item the conference information in the page header(s).
\end{itemize}

Rights information is unique to the work; if you are preparing several
works for an event, make sure to use the correct set of commands with
each of the works.

The ACM Reference Format text is required for all articles over one
page in length, and is optional for one-page articles (abstracts).

\section{CCS Concepts and User-Defined Keywords}

Two elements of the ``acmart'' document class provide powerful
taxonomic tools for you to help readers find your work in an online
search.

The ACM Computing Classification System ---
\url{https://www.acm.org/publications/class-2012} --- is a set of
classifiers and concepts that describe the computing
discipline. Authors can select entries from this classification
system, via \url{https://dl.acm.org/ccs/ccs.cfm}, and generate the
commands to be included in the \LaTeX\ source.

User-defined keywords are a comma-separated list of words and phrases
of the authors' choosing, providing a more flexible way of describing
the research being presented.

CCS concepts and user-defined keywords are required for for all
articles over two pages in length, and are optional for one- and
two-page articles (or abstracts).

\section{Sectioning Commands}

Your work should use standard \LaTeX\ sectioning commands:
\verb|section|, \verb|subsection|, \verb|subsubsection|, and
\verb|paragraph|. They should be numbered; do not remove the numbering
from the commands.

Simulating a sectioning command by setting the first word or words of
a paragraph in boldface or italicized text is {\bfseries not allowed.}

\section{Tables}

The ``\verb|acmart|'' document class includes the ``\verb|booktabs|''
package --- \url{https://ctan.org/pkg/booktabs} --- for preparing
high-quality tables.

Table captions are placed {\itshape above} the table.

Because tables cannot be split across pages, the best placement for
them is typically the top of the page nearest their initial cite.  To
ensure this proper ``floating'' placement of tables, use the
environment \textbf{table} to enclose the table's contents and the
table caption.  The contents of the table itself must go in the
\textbf{tabular} environment, to be aligned properly in rows and
columns, with the desired horizontal and vertical rules.  Again,
detailed instructions on \textbf{tabular} material are found in the
\textit{\LaTeX\ User's Guide}.

Immediately following this sentence is the point at which
Table~\ref{tab:freq} is included in the input file; compare the
placement of the table here with the table in the printed output of
this document.

\begin{table}
  \caption{Frequency of Special Characters}
  \label{tab:freq}
  \begin{tabular}{ccl}
    \toprule
    Non-English or Math&Frequency&Comments\\
    \midrule
    \O & 1 in 1,000& For Swedish names\\
    $\pi$ & 1 in 5& Common in math\\
    \$ & 4 in 5 & Used in business\\
    $\Psi^2_1$ & 1 in 40,000& Unexplained usage\\
  \bottomrule
\end{tabular}
\end{table}

To set a wider table, which takes up the whole width of the page's
live area, use the environment \textbf{table*} to enclose the table's
contents and the table caption.  As with a single-column table, this
wide table will ``float'' to a location deemed more
desirable. Immediately following this sentence is the point at which
Table~\ref{tab:commands} is included in the input file; again, it is
instructive to compare the placement of the table here with the table
in the printed output of this document.

\begin{table*}
  \caption{Some Typical Commands}
  \label{tab:commands}
  \begin{tabular}{ccl}
    \toprule
    Command &A Number & Comments\\
    \midrule
    \texttt{{\char'134}author} & 100& Author \\
    \texttt{{\char'134}table}& 300 & For tables\\
    \texttt{{\char'134}table*}& 400& For wider tables\\
    \bottomrule
  \end{tabular}
\end{table*}

Always use midrule to separate table header rows from data rows, and
use it only for this purpose. This enables assistive technologies to
recognise table headers and support their users in navigating tables
more easily.

\section{Math Equations}
You may want to display math equations in three distinct styles:
inline, numbered or non-numbered display.  Each of the three are
discussed in the next sections.

\subsection{Inline (In-text) Equations}
A formula that appears in the running text is called an inline or
in-text formula.  It is produced by the \textbf{math} environment,
which can be invoked with the usual
\texttt{{\char'134}begin\,\ldots{\char'134}end} construction or with
the short form \texttt{\$\,\ldots\$}. You can use any of the symbols
and structures, from $\alpha$ to $\omega$, available in
\LaTeX~\cite{Lamport:LaTeX}; this section will simply show a few
examples of in-text equations in context. Notice how this equation:
\begin{math}
  \lim_{n\rightarrow \infty}x=0
\end{math},
set here in in-line math style, looks slightly different when
set in display style.  (See next section).

\subsection{Display Equations}
A numbered display equation---one set off by vertical space from the
text and centered horizontally---is produced by the \textbf{equation}
environment. An unnumbered display equation is produced by the
\textbf{displaymath} environment.

Again, in either environment, you can use any of the symbols and
structures available in \LaTeX\@; this section will just give a couple
of examples of display equations in context.  First, consider the
equation, shown as an inline equation above:
\begin{equation}
  \lim_{n\rightarrow \infty}x=0
\end{equation}
Notice how it is formatted somewhat differently in
the \textbf{displaymath}
environment.  Now, we'll enter an unnumbered equation:
\begin{displaymath}
  \sum_{i=0}^{\infty} x + 1
\end{displaymath}
and follow it with another numbered equation:
\begin{equation}
  \sum_{i=0}^{\infty}x_i=\int_{0}^{\pi+2} f
\end{equation}
just to demonstrate \LaTeX's able handling of numbering.

\section{Figures}

The ``\verb|figure|'' environment should be used for figures. One or
more images can be placed within a figure. If your figure contains
third-party material, you must clearly identify it as such, as shown
in the example below.

\begin{figure}[h]
  \centering
    \fbox{\rule{0pt}{2.5in} \rule{0.9\linewidth}{0pt}}
  \caption{Example of caption}
\end{figure}

Your figures should contain a caption which describes the figure to
the reader.

Figure captions are placed {\itshape below} the figure.

Every figure should also have a figure description unless it is purely
decorative. These descriptions convey what’s in the image to someone
who cannot see it. They are also used by search engine crawlers for
indexing images, and when images cannot be loaded.

A figure description must be unformatted plain text less than 2000
characters long (including spaces).  {\bfseries Figure descriptions
  should not repeat the figure caption – their purpose is to capture
  important information that is not already provided in the caption or
  the main text of the paper.} For figures that convey important and
complex new information, a short text description may not be
adequate. More complex alternative descriptions can be placed in an
appendix and referenced in a short figure description. For example,
provide a data table capturing the information in a bar chart, or a
structured list representing a graph.  For additional information
regarding how best to write figure descriptions and why doing this is
so important, please see
\url{https://www.acm.org/publications/taps/describing-figures/}.

\subsection{The ``Teaser Figure''}

A ``teaser figure'' is an image, or set of images in one figure, that
are placed after all author and affiliation information, and before
the body of the article, spanning the page. If you wish to have such a
figure in your article, place the command immediately before the
\verb|\maketitle| command:
\begin{verbatim}
  \begin{teaserfigure}
    \includegraphics[width=\textwidth]{sampleteaser}
    \caption{figure caption}
    \Description{figure description}
  \end{teaserfigure}
\end{verbatim}

\section{Citations and Bibliographies}

The use of \BibTeX\ for the preparation and formatting of one's
references is strongly recommended. Authors' names should be complete
--- use full first names (``Donald E. Knuth'') not initials
(``D. E. Knuth'') --- and the salient identifying features of a
reference should be included: title, year, volume, number, pages,
article DOI, etc.

The bibliography is included in your source document with these two
commands, placed just before the \verb|\end{document}| command:
\begin{verbatim}
  \bibliographystyle{ACM-Reference-Format}
  \bibliography{bibfile}
\end{verbatim}
where ``\verb|bibfile|'' is the name, without the ``\verb|.bib|''
suffix, of the \BibTeX\ file.

Citations and references are numbered by default. A small number of
ACM publications have citations and references formatted in the
``author year'' style; for these exceptions, please include this
command in the {\bfseries preamble} (before the command
``\verb|\begin{document}|'') of your \LaTeX\ source:
\begin{verbatim}
  \citestyle{acmauthoryear}
\end{verbatim}

  Some examples.  A paginated journal article \cite{Abril07}, an
  enumerated journal article \cite{Cohen07}, a reference to an entire
  issue \cite{JCohen96}, a monograph (whole book) \cite{Kosiur01}, a
  monograph/whole book in a series (see 2a in spec. document)
  \cite{Harel79}, a divisible-book such as an anthology or compilation
  \cite{Editor00} followed by the same example, however we only output
  the series if the volume number is given \cite{Editor00a} (so
  Editor00a's series should NOT be present since it has no vol. no.),
  a chapter in a divisible book \cite{Spector90}, a chapter in a
  divisible book in a series \cite{Douglass98}, a multi-volume work as
  book \cite{Knuth97}, a couple of articles in a proceedings (of a
  conference, symposium, workshop for example) (paginated proceedings
  article) \cite{Andler79, Hagerup1993}, a proceedings article with
  all possible elements \cite{Smith10}, an example of an enumerated
  proceedings article \cite{VanGundy07}, an informally published work
  \cite{Harel78}, a couple of preprints \cite{Bornmann2019,
    AnzarootPBM14}, a doctoral dissertation \cite{Clarkson85}, a
  master's thesis: \cite{anisi03}, an online document / world wide web
  resource \cite{Thornburg01, Ablamowicz07, Poker06}, a video game
  (Case 1) \cite{Obama08} and (Case 2) \cite{Novak03} and \cite{Lee05}
  and (Case 3) a patent \cite{JoeScientist001}, work accepted for
  publication \cite{rous08}, 'YYYYb'-test for prolific author
  \cite{SaeediMEJ10} and \cite{SaeediJETC10}. Other cites might
  contain 'duplicate' DOI and URLs (some SIAM articles)
  \cite{Kirschmer:2010:AEI:1958016.1958018}. Boris / Barbara Beeton:
  multi-volume works as books \cite{MR781536} and \cite{MR781537}. A
  couple of citations with DOIs:
  \cite{2004:ITE:1009386.1010128,Kirschmer:2010:AEI:1958016.1958018}. Online
  citations: \cite{TUGInstmem, Thornburg01, CTANacmart}. Artifacts:
  \cite{R} and \cite{UMassCitations}.

\section{Acknowledgments}

Identification of funding sources and other support, and thanks to
individuals and groups that assisted in the research and the
preparation of the work should be included in an acknowledgment
section, which is placed just before the reference section in your
document.

This section has a special environment:
\begin{verbatim}
  \begin{acks}
  ...
  \end{acks}
\end{verbatim}
so that the information contained therein can be more easily collected
during the article metadata extraction phase, and to ensure
consistency in the spelling of the section heading.

Authors should not prepare this section as a numbered or unnumbered {\verb|\section|}; please use the ``{\verb|acks|}'' environment.

\section{Appendices}

If your work needs an appendix, add it before the
``\verb|\end{document}|'' command at the conclusion of your source
document.

Start the appendix with the ``\verb|appendix|'' command:
\begin{verbatim}
  \appendix
\end{verbatim}
and note that in the appendix, sections are lettered, not
numbered. This document has two appendices, demonstrating the section
and subsection identification method.

\section{Multi-language papers}

Papers may be written in languages other than English or include
titles, subtitles, keywords and abstracts in different languages (as a
rule, a paper in a language other than English should include an
English title and an English abstract).  Use \verb|language=...| for
every language used in the paper.  The last language indicated is the
main language of the paper.  For example, a French paper with
additional titles and abstracts in English and German may start with
the following command
\begin{verbatim}
\documentclass[sigconf, language=english, language=german,
               language=french]{acmart}
\end{verbatim}

The title, subtitle, keywords and abstract will be typeset in the main
language of the paper.  The commands \verb|\translatedXXX|, \verb|XXX|
begin title, subtitle and keywords, can be used to set these elements
in the other languages.  The environment \verb|translatedabstract| is
used to set the translation of the abstract.  These commands and
environment have a mandatory first argument: the language of the
second argument.  See \verb|sample-sigconf-i13n.tex| file for examples
of their usage.

\section{SIGCHI Extended Abstracts}

The ``\verb|sigchi-a|'' template style (available only in \LaTeX\ and
not in Word) produces a landscape-orientation formatted article, with
a wide left margin. Three environments are available for use with the
``\verb|sigchi-a|'' template style, and produce formatted output in
the margin:
\begin{itemize}
\item {\verb|sidebar|}:  Place formatted text in the margin.
\item {\verb|marginfigure|}: Place a figure in the margin.
\item {\verb|margintable|}: Place a table in the margin.
\end{itemize}


\bibliographystyle{ACM-Reference-Format}
\bibliography{sample-base}








